\documentclass[10pt,twocolumn,letterpaper]{article}

\usepackage{iccv}
\usepackage{times}
\usepackage{epsfig}
\usepackage{graphicx}
\usepackage{amsmath}
\usepackage{amssymb}

\usepackage{epstopdf}
\DeclareGraphicsExtensions{.png,.pdf}

\usepackage{booktabs}

\newcommand{\comment}[1]{}

\usepackage{siunitx}

\usepackage{float}

\usepackage[breaklinks=true,bookmarks=false]{hyperref}

\iccvfinalcopy 

\def\iccvPaperID{****} 

\ificcvfinal\fi

\begin{document}

\title{Learning Dense Wide Baseline Stereo Matching for People}

\author{Akin Caliskan$^{1}$, Armin Mustafa$^{1}$, Evren Imre$^{2}$, Adrian Hilton$^{1}$\\
$^{1}${Center for Vision, Speech and Signal Processing}\\
University of Surrey, UK \\
$^{2}$ Vicon Motion Systems Ltd. \\
{\tt\small $^{1}$\{a.caliskan, a.mustafa, a.hilton\}@surrey.ac.uk, $^{2}$evren.imre@vicon.com }
}

\maketitle
\ificcvfinal\thispagestyle{empty}\fi

\begin{abstract}
    Existing methods for stereo work on narrow baseline image pairs giving limited performance between wide baseline views. This paper proposes a framework to learn and estimate dense stereo for people from wide baseline image pairs. A synthetic people stereo patch dataset \textit(S2P2) is introduced to learn wide baseline dense stereo matching for people.
   The proposed framework not only learns human specific features from synthetic data but also exploits pooling layer and data augmentation to adapt to real data. The network learns from the human specific stereo patches from the proposed dataset for wide-baseline stereo estimation. 
   In addition to patch match learning, a stereo constraint is introduced in the framework to solve wide baseline stereo reconstruction of humans. Quantitative and qualitative performance evaluation against state-of-the-art methods of proposed method demonstrates improved wide baseline stereo reconstruction on challenging datasets. We show that it is possible to learn stereo matching from synthetic people dataset and improve performance on real datasets for stereo reconstruction of people from narrow and wide baseline stereo data.
\end{abstract}


\section{Introduction}


Recent developments in augmented reality/virtual reality and autonomous driving has led to a need for high-quality 3D content, especially for humans. However, existing scanning technologies require advanced camera setups, and controlled studio capture environments, which are complex and costly solutions. To address the need for democratization of high-quality 3D content, we propose dense stereo reconstruction for humans from wide baseline image pairs

\begin{figure}[t]
    \includegraphics[width=\columnwidth]{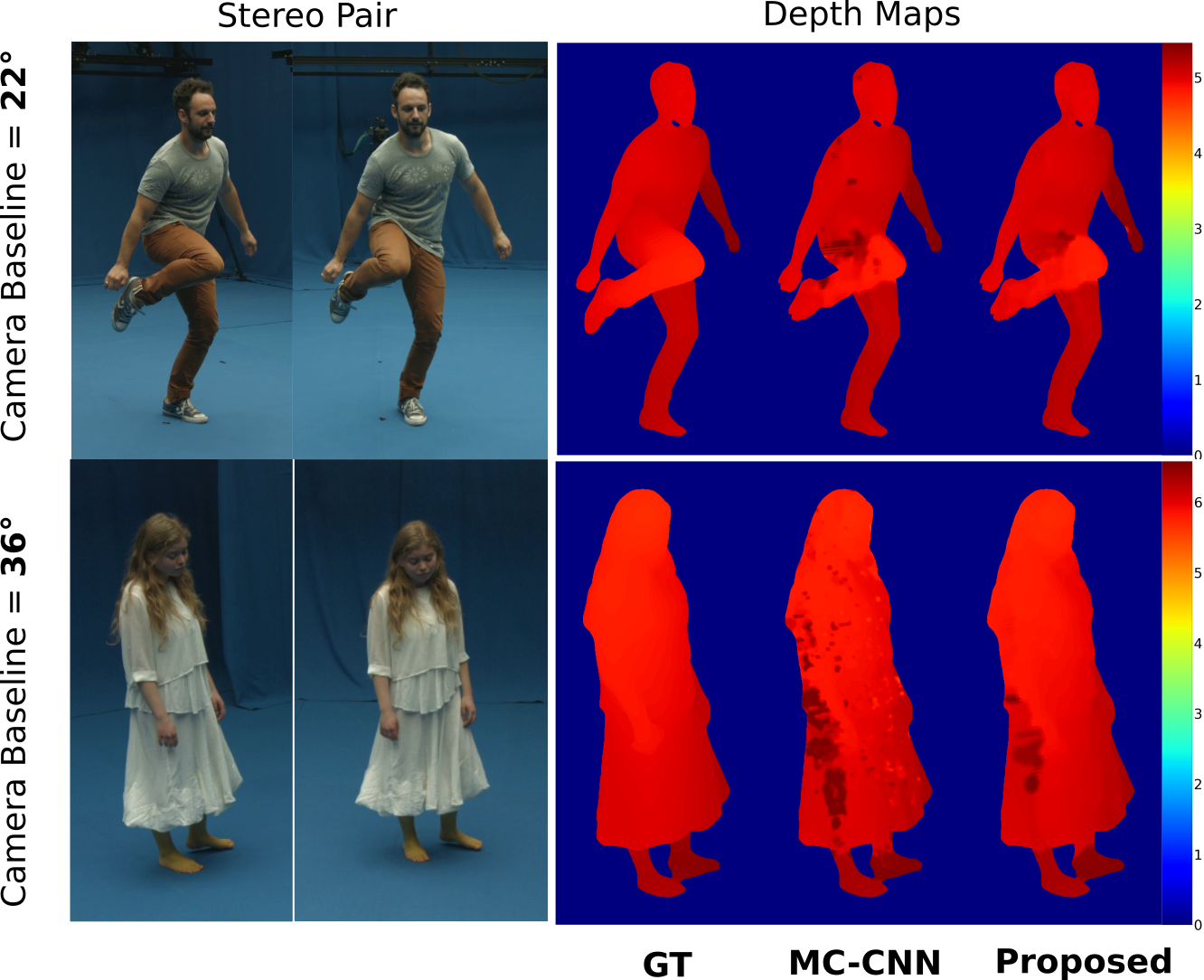}
  \caption{This figure shows the wide baseline stereo input and depth map estimation of our method with state of the art patch based stereo disparity estimation MC-CNN \cite{zbontar2016stereo} compared to ground truth depth map. \newline }
\end{figure}\label{Sota_compare}


Existing dense stereo reconstruction methods are broadly divided in two groups; narrow baseline and wide baseline stereo. For narrow baseline stereo, it is possible to estimate pixel matches by using conventional \cite{hirschmuller2007stereo,scharstein2002taxonomy} or learning based methods \cite{zagoruyko2015learning,zbontar2016stereo,kendall2017end,chabra2019stereodrnet}. Recently learning based methods have gained attention by outperforming conventional methods, as illustrated in the benchmark \cite{Menze2015CVPRKitti}.
However, for wide baseline stereo, the research has focused on conventional methods \cite{tola2009daisy,hu2012quantitative}, and data driven learning based approaches are still an open research question because of the lack of training data.
Results from \cite{leroy2017multi, richardt2016dense, mustafa2015general} demonstrate that conventional wide baseline stereo methods have limitation on finding accurate matching for the human body surface. Inspired by the narrow-baseline learning based approaches and need for human specific wide baseline stereo matching, we propose a framework to estimate wide baseline dense stereo matching for people.
We exploit a Siamese architecture \cite{siamese_network} and fully connected network to learn stereo matching (Section \ref{M_CNN}). However existing datasets for learning stereo matching are designed for narrow baseline images with fixed relative camera locations \cite{mayer2018makes,Menze2015CVPRKitti,VirtualKitti}. 
In this work we introduce a synthetic human specific wide baseline stereo dataset to overcome 
the limitations of existing datasets. To refine the stereo matching performance we also propose the use of constrained stereo search using a semantic mask which is demonstrated to further refine the stereo matching performance.


Recently, the Mannequin Challenge dataset \cite{li2019learning} has been introduced to learn monocular depth estimation for humans from a dataset of frozen people in the scene. However, this dataset does not address the problem of learning stereo matching from pairs of images.
Hence, there is a need for a new dataset to perform stereo correspondence with significant change in appearance between views due to the high surcease shape of humans which includes dynamic non-rigid motion and loose clothing. Capturing a high-frequency human body with accurate ground-truth shape requires advanced scanning system which is expensive and is not easily accessible.
Hence, we propose to generate stereo patch dataset for people (\textit{S2P2}) from synthetic 3D human models with realistic textures. This dataset is used to train the stereo matching network to learn features to compare stereo image patches on the human body surface. 
Commonly networks trained with synthetic data do not perform well on real datasets due to the problem of domain shift \cite{mayer2018makes}. Methods have tried to address this problem in the literature for different applications such as semantic segmentation \cite{sankaranarayanan2018learning}, stereo reconstruction and optical flow estimation \cite{mayer2018makes} and scene understanding \cite{sakaridis2018model}. These methods increase the variation in dataset by augmentation of training data with random spatial operations \cite{bloice2017augmentor} or by creating realistic data. We have exploited these ideas by using realistic textures and applying augmentation to generated patches (Section \ref{M_datagen}).
Another problem introduced by the domain gap is \textit{scale diversity}. The scale of objects such as people in the image is unknown and potentially limiting the performance of a model trained on synthetic dataset. \cite{pang2018zoom} proposes to use stereo pairs in various sizes to generalise training for  scale diversity. Inspired by this work we extract features at different scales and combine them to get the final matching cost. We demonstrate the accuracy of the proposed stereo reconstruction on cluttered real world dataset of people in the experiments. A comprehensive performance evaluation is performed to evaluate our method with ground-truth 3D reconstruction of dynamic shape from state-of-the-art from multiple views studio performance capture. Comparison of our method with baseline methods for stereo matching shows the superior performance of features learned from the \textit{S2P2} synthetic human dataset on wide baseline dense dynamic human stereo reconstruction. Our contributions are: 
\vspace{-5pt}
\begin{itemize}
  \item Introduction of the first learning based framework to estimate dense wide baseline stereo for people.
  \vspace{-5pt}
  \item A large scale, synthetic stereo patch dataset for people with realistic textures for both narrow and wide camera baseline stereo.
  \vspace{-5pt}
  \item Augmentation of data and matching across multiple scales to make proposed method robust to problem of domain shift and scale.
  \vspace{-5pt}
  \item Refinement of learnt human stereo matching using a semantic human mask for improved stereo reconstruction.
\end{itemize}


\section{Related Work}

\paragraph{Dynamic Human Stereo Reconstruction:} 

Existing methods for stereo reconstruction of dynamic scenes estimate correspondences between image pairs to obtain accurate surface reconstruction \cite{richardt2016dense,leroy2017multi} for wide baseline images. 
Daisy \cite{tola2009daisy} and Normalized Cross Correlation (NCC) \cite{hu2012quantitative} uses gradient of local patch's around pixels to compute descriptor or pixel colour distribution of local patch \cite{hu2012quantitative} to measure patch correlation for dense wide baseline matching.
In previous approaches, computation of a patch similarity measure is used as a photo-metric loss term in the objective function of optimization schema which exploits other priors, such as optical flow, edges or foreground/background segmentation \cite{mustafa2015general,richardt2016dense,leroy2017multi}. In other words, dynamic wide baseline stereo reconstruction has not been considered as an individual solution. 

Recently, learning based approaches for stereo matching have gained attention  \cite{zagoruyko2015learning,zbontar2016stereo,kendall2017end,chabra2019stereodrnet} for stereo disparity estimation. However, these are trained on general scenes and are limited to narrow baseline stereo matching. To the best of our knowledge, human specific wide baseline stereo has not been addressed with learning based approaches before. Previous work \cite{leroy2018shape} trains a multi-view patch similarity network for performance capture using the DTU general object dataset \cite{aanaes2016large}.
This paper addresses this gap in the literature, by proposing dense stereo reconstruction from wide baseline image pairs and learning to perform stereo matching using a new synthetic people stereo patch dataset (\textit{S2P2}).

\vspace{-15pt}

\paragraph{Learning Depth from Synthetic Data}

Recently, usage of synthetic data to train neural networks for depth estimation has gained attention. One of the first synthetic data-set proposed is \cite{mayer2018makes}. This data is used to train a network for narrow baseline stereo disparity and optical flow estimation.
Another work from \cite{VirtualKitti} generates the virtual version of Kitti data-set \cite{Menze2015CVPRKitti}. Virtual Kitti includes additional annotations like segmentation, depth estimation and 3D object tracking. They demonstrate that training on synthetic data and using learned model on real data is possible.
Varol \textit{et al.} \cite{varol17_surreal} proposed a synthetic human dataset for monocular model based human segmentation and depth estimation. However, synthetic data trained models suffer from limitations on real world images in high-frequency depth estimation of the human body \cite{saito2019pifu}. \cite{huang2018deep} introduced another synthetic human dataset to train multi-view surface estimation network. We propose a large scale stereo patch dataset \textit(S2P2) for people to train a network for wide baseline dense stereo matching across difference scales. To the best of our knowledge, the proposed \textit(S2P2) dataset is the first to learn stereo matching for people. This dataset can be used for both narrow and wide baseline stereo estimation.

\begin{figure*}[t]
  \includegraphics[width=\textwidth]{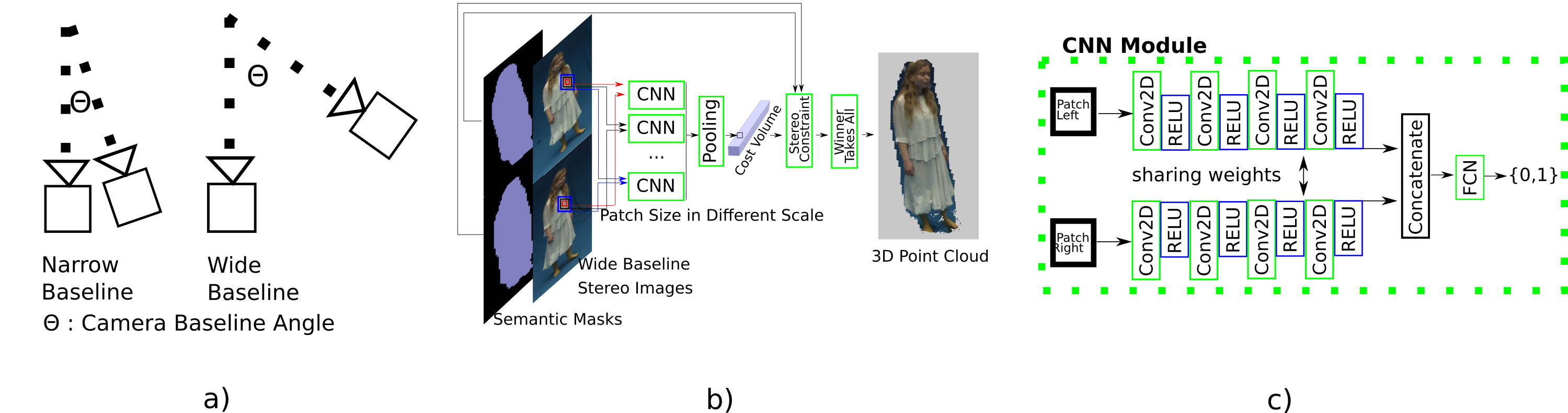}
  \caption{(a) Definition of camera baseline angle, $\theta$ and difference between narrow and wide camera baselines.(b) The proposed stereo reconstruction method.(c) CNN module used for patch match learning part of the proposed method.}
\end{figure*}\label{method_overview}

\begin{figure}[t]
  \includegraphics[width=\columnwidth]{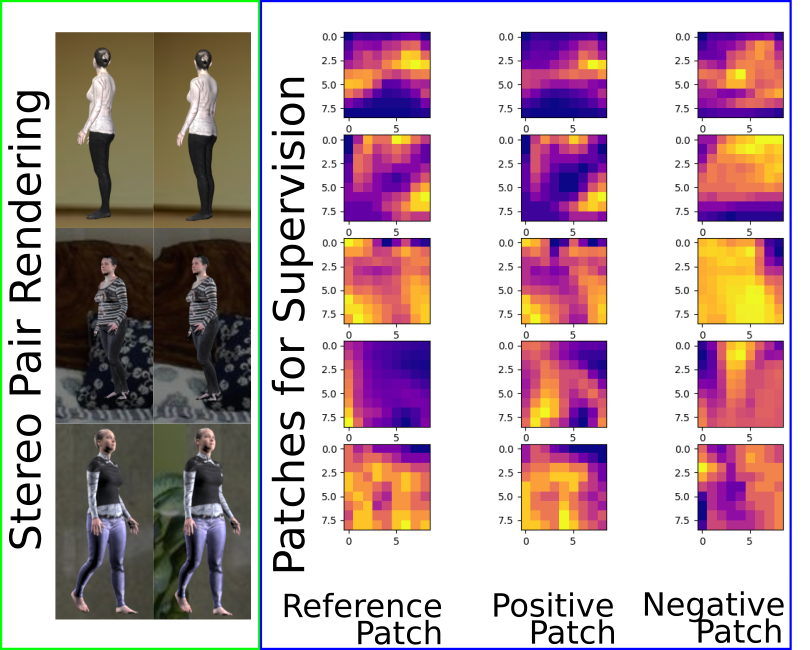}
  \caption{\textbf{Data Generation Pipeline:} SMPL human model \cite{SMPL_2015} is generated with random shape, texture and given 3D pose parameters. Cameras are placed according to real studio calibration in different baseline configuration. [\textit{Left}] Rendering of human models onto camera planes with different background and 3D pose. [\textit{Right}] Positive and negative patches are generated from these images. For details, please refer to text. \newline }
\end{figure}\label{fig_M_data_gen}


\section{Method}

\vspace{-5pt}

The main motivation of this work is to estimate 3D reconstruction of humans in dynamic scenes from wide baseline stereo camera pairs. Note that, the difference between narrow ( $\theta <= 5^{\circ}$ ) and wide baseline  ( $ 15^{\circ} < \theta < 45^{\circ}$ ) cameras is illustrated in Figure \ref{method_overview} - (a).
We propose a supervised learning based framework which first learns stereo matching from a new synthetic human specific dataset \textit{S2P2} for wide-baseline cameras followed by stereo reconstruction refinement using semantic human constraint, an overview is illustrated in Figure \ref{method_overview} - (b). 
Variation of human body surface for example folded clothing, hair, face details, makes it challenging to extract reliable stereo reconstruction from wide baseline image pairs.
Given a wide baseline stereo pair of images of a person, we aim to obtain per-pixel dense correspondence for stereo reconstruction. 
The stereo pair of images are fed into a CNN module, which is trained on a human specific dataset to obtain the matching cost for each pixel. This generates a cost volume which is refined using a semantic stereo constraint to obtain the final depth map. 
In the following sections, the patch match learning architecture (Section \ref{M_CNN}), data generation pipeline (Section \ref{M_datagen}), the method to solve domain shift from synthetic to real data (Section \ref{M_dom}) and semantic stereo constraint (Section \ref{M_ste_cons}) are explained in detail.


\subsection{Learning Wide Baseline Stereo Matching} \label{M_CNN}

The overall CNN module for learning stereo matching is illustrated in Figure \ref{method_overview} - (c). We use a Siamese network architecture \cite{siamese_network} as the backbone, which has received a lot of attention lately for various applications including patch based binary classification \cite{zbontar2016stereo}, and patch based tracking \cite{bertinetto2016fully}. Siamese network is suitable for the proposed application because it allows training of stereo matching between a pair of left and right image patches. Methods used Siamese network as feature extraction module in patch based narrow baseline stereo matching \cite{luo2016efficient,zbontar2016stereo}. 
The network consists of four consecutive 2D convolution layers and RELU (Rectified Linear Units) after each convolution layer. As illustrated in Figure \ref{method_overview} - (c), the computed feature vectors are fed into a fully-connected network (FCN) to estimate the similarity score between patches, i.e. classification module. The details of CNN module is provided in the supplementary file. Since we are solving a binary classification problem, we use binary cross entropy loss \cite{murphy2012machine} to train our network. During the training stage, we use a balanced number of positive and negative patches extracted from the \textit{S2P2} dataset (Sec. \ref{M_datagen}). 

In the implementation stage, multi resolution patches are extracted for each pixel followed by resizing the patches to a fixed patch size that the network is trained with. Patches are processed through the network, and matching cost is computed for each patch.
Individually generated cost volumes are fed into the pooling stage as illustrated in Figure \ref{method_overview}, where the resultant matching cost is computed from the similarity scores for each pixel pairs.
In the pooling stage, the matching cost from different patch sizes are gathered and the average value is assigned as a final cost. We evaluate the effect of pooling by comparing the results with or without pooling in the Experiment section - Table \ref{tab:table2_scale_diversity}.

We compute the cost volume for both left and right camera views respectively and a winner takes all method is applied to each of the views to compute the final disparity values. 
In contrast to the conventional stereo pipelines \cite{scharstein2002taxonomy, zbontar2016stereo} which require heavy regularization steps for post-processing like Semi-Global Matching \cite{hirschmuller2007stereo} and Bilateral filtering \cite{scharstein2002taxonomy}, we perform a simple post-processing to remove the occlusions on the estimated disparity maps to improve stereo from wide baseline image pairs. 


\subsection{Synthetic People Stereo Patch Dataset} \label{M_datagen}
Existing datasets in the literature are limited to narrow-baseline general scenes. We address this gap in the literature by proposing a data generation framework for supervised wide baseline stereo matching learning for people, illustrated in Figure \ref{fig_M_data_gen}. 
We generate the dataset by using the blender 3D modelling\footnote{https://www.blender.org}. Parametric 3D SMPL \cite{SMPL_2015} human models are generated based on 3D pose estimation from real humans with random shape parameters, CMU MoCap Dataset \cite{CMU_MoCap}. Then realistic textures are rendered on the generated models. Up to this point, model generation is inspired by the Surreal dataset \cite{varol17_surreal}.

To add varied backgrounds to each image a 3D plane is placed behind the person model and background scene images are randomly selected from Places Database \cite{zhou2014learning}, which consists of high variation of indoor and outdoor places with different configurations. 
Camera locations and orientations are replicated from real studio capture setups, and the baseline between cameras is varied from narrow ($~5^{\circ}$) to wide (up to $45^{\circ}$ degrees). The generated scene is then rendered into camera views with random lighting settings. For training purposes, we generate patches from non-occluded regions of the human body surface. 

Proposed network structure requires positive and negative patches. Positive patches are generated from projection of 3D points into stereo views and negative patches are $\psi$ pixels away from positive patches along the Epipolar line. During training data generation, patch size is fixed to 9x9, and $\psi$ value is randomly selected from interval $[4,11]$. Reference patch with positive and negative pairs are augmented \cite{bloice2017augmentor} in spatial and spectral domains which includes random cropping, flipping, transformation, and contrast variation. 
This dataset along with the data generation framework is available for public use \footnote{https://akcalakcal.github.io/Learning-Dense-Wide-Baseline-Stereo-Matching-for-People/} and further details of the dataset are given in supplementary material. 

\vspace{-10pt}

\paragraph{Data Augmentation and Scale Invariance:} \label{M_dom}
Learning from the synthetic dataset and testing on real images has recently gained attention in the literature for different applications \cite{mayer2018makes,VirtualKitti, varol17_surreal,rad2018feature}. The common problem is domain adaptation which directly affects the learning from synthetic to real imagery. In our work, we generate \textit{S2P2} dataset from a wide variety of camera positions and realistically textured human models. We add patch augmentation, explained previously, to increase the robustness of stereo correspondence for real data.
However real data can be observed with different input scale than synthetic data, which results in stereo correspondence defects, called \textit{scale diversity} \cite{pang2018zoom}. Since the scale of real data is unknown, we look for the consistency of accurate matches for different patch sizes before computation of the final cost volume in the \textit{pooling stage}, illustrated in Figure \ref{method_overview} - (b). To address this, we take the average of matching cost values that are computed with the trained network for every pixel. This multi-scale patch size approach is analyzed for real data and Table \ref{tab:table2_scale_diversity} shows the performance improvement in the reconstruction accuracy. 

\begin{figure}[ht]
  \includegraphics[width=\columnwidth]{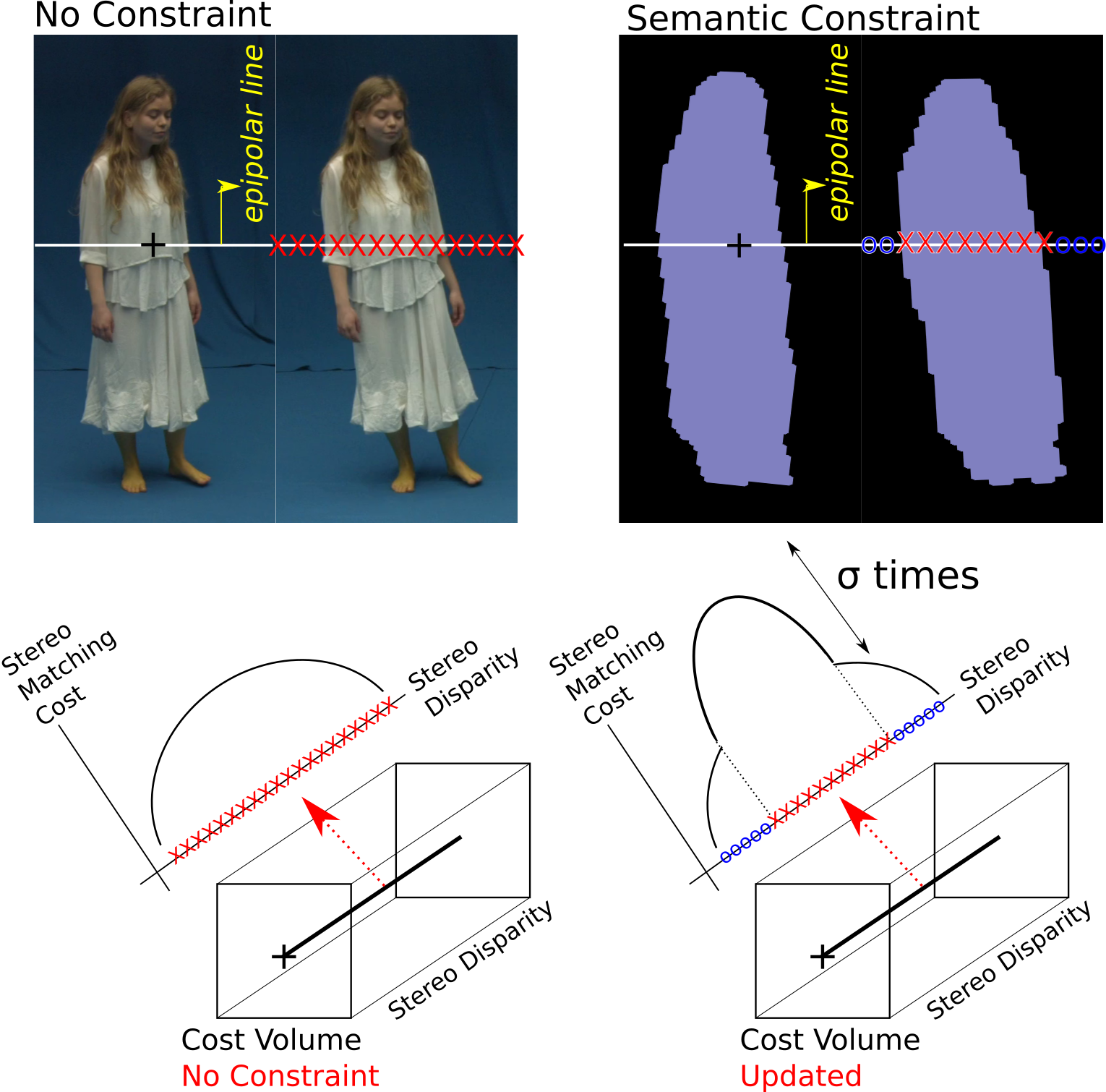}
  \caption{Semantic mask based stereo constraint for wide baseline stereo on Acting dataset \cite{CVSSP_3D}.}
  \label{fig_M_stereo_const}
\end{figure}

\subsection{Semantic Stereo Constraints} \label{M_ste_cons}
To further refine the learnt stereo matching we introduce a semantic stereo constraint for stereo matching on people leveraging recent advances in semantic segmentation. Stereo matching requires reliable per-pixel correspondences in between image pairs. With a given calibration, patch match methods rectify the images to find correspondences along the epipolar line by comparing the pixel similarities. However due to drastic view variation in wide baseline stereo pairs, patch match methods fail to find reliable correspondences. 
Previous studies on wide baseline human performance capture methods \cite{mustafa2015general,starck2007} either use initial sparse reconstruction or visual hulls generated from multi cameras to limit the stereo search space. Other methods for wide baseline semantic reconstruction exploit semantic segmentation constraints to improve the multi-view stereo \cite{mustafa2017semantically}. In this study, we propose to exploit semantic masking in the stereo matching framework to limit the search region along the Epipolar line to decrease the number of wrong matches from only two camera views. However errors in semantic segmentation do not adversely affect the accuracy of the reconstruction, unlike previous method.

We use DeepLabv3+ \cite{deeplabv3plus2018} to obtain the semantic masks. The correspondence search algorithm for two stereo rectified images and corresponding semantic masks is illustrated in Figure \ref{fig_M_stereo_const}. 
Without constraint, a pixel in the left image is compared with all the pixels in the corresponding right image. However, with the semantic constraint we search for the corresponding pixel within the semantic region along the Epipolar line reducing the ambiguity and run-time complexity.
The cost volume in Figure \ref{fig_M_stereo_const} is processed with the semantic constraint such that for pixels in the masked region, the cost value is weighted by a coefficient, $\sigma = 10$. This suppress other pixels for matching.

\vspace{-5pt}

\section{Experiments}

\begin{figure}[t]
 \centering
  \includegraphics[width=6cm]{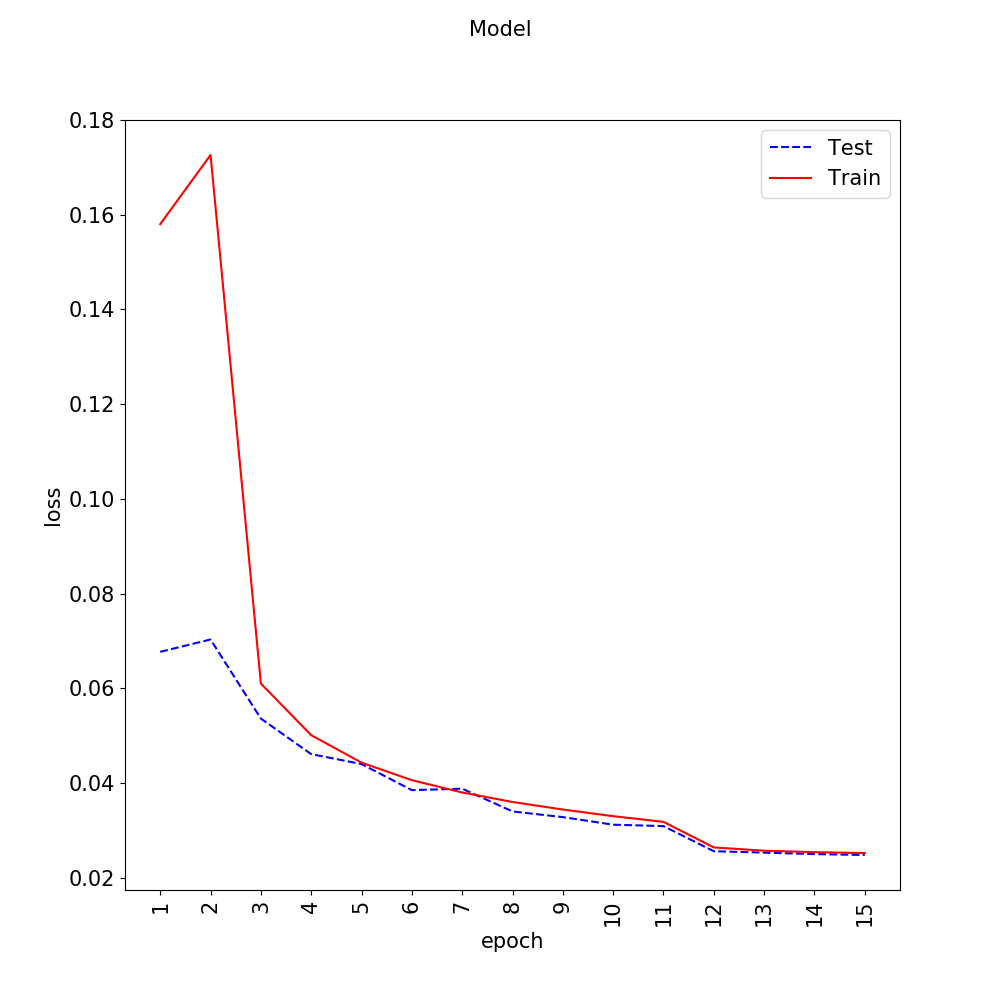}
  \caption{Variation of loss during training.}
  \label{Exp_tr_loss}
\end{figure}

We answer the following questions in experiments:
\vspace{-10pt}
\begin{itemize}
    \item Does learning wide baseline stereo matching from people dataset result in better matching for image pairs of people that existing approaches which learn from non-human stereo dataset and conventional methods? 
    \vspace{-10pt}
    \item Does the proposed solution to domain shift with patch augmentation and scale diversity, improve the reconstruction results for real datasets with humans? 
    \vspace{-10pt}
    \item Does the proposed semantic human stereo constraint improve the stereo reconstruction results? 
\end{itemize}
\paragraph{Implementation and Training Details}
The network architecture is implemented in PyTorch \cite{pytorch_citation} framework on a single NVIDIA GeForce 1080 Ti GPU with 12 GB memory. As described in Section \ref{M_CNN}, we train our model from scratch. The learning rate is initialized at $3\times10^{-3}$ with a 10 times decrease at every 10 epochs. Training is performed for 15 epochs, and momentum and weight decay are set to $0.9$ and $0.0001$, respectively. The entire network is learned with stochastic gradient descent optimization with binary cross entropy loss function. The network weights are randomly initialized with balanced number of positive and negative patches with a total of 14 million patches. Variation of training loss versus epochs is illustrated in Figure \ref{Exp_tr_loss}.

\begingroup
\begin{table}[b]
\resizebox{\columnwidth}{!}{
\begin{tabular}{llll}
\\  \hline \\ 
    Dataset &  Camera Baseline & \# of Cameras & Length of Sequence 
    \\
    & & & (\# of Frames)
    \\    \hline
 Acting         & \{$24^{\circ}$,$36^{\circ}$,$48^{\circ}$\} & 15 & 3420  \\
 TV Presenter   & \{$22^{\circ}$,$44^{\circ}$,$66^{\circ}$\} & 16  & 3600  \\
 Dancing        & \{$22^{\circ}$,$44^{\circ}$,$66^{\circ}$\} & 16 & 420   \\
 Juggler        & \{$22^{\circ}$,$44^{\circ}$,$66^{\circ}$\} & 8 & 800   \\

\end{tabular}}
\caption{Real World of People Datasets.} 
\label{tab:table_real_dataset}
\end{table}
\endgroup

\begingroup
\begin{table*}[t]
\tiny
\resizebox{\textwidth}{!}{\begin{tabular}{lllllllll}
\\    \hline
 & \multicolumn{4}{c}{Camera Baseline $\approx$ ${{20}^{\circ}}$} & \multicolumn{4}{c}{Camera Baseline $\approx$ ${{~40}^{\circ}}$}    \\ 
 \cline{2-5} \cline{6-9} \\
    Method &  Abs Rel   & Squ Rel & RMSE & $\text{RMSE}_{log}$ &  Abs Rel & Squ Rel & RMSE & $\text{RMSE}_{log}$
    \\    \hline
     \textbf{Dataset:Acting}
     \\
 NCC \cite{hu2012quantitative}  & 3.40 & 3.18 & 44.3 & 7.89  & 5.21 & 3.52 & 46.6 & 8.10         \\       
 Daisy \cite{tola2009daisy}   & 1.77 & 0.92 & 24.0 & 3.53  & 2.05 & 0.95 & 24.1 & 3.70         \\         
 MC-CNN \cite{zbontar2016stereo}   & 1.27 & 0.11 & 8.56 & 1.36 & 1.42 & 0.43 & 16.1 & 2.57   \\
  Ours & \textbf{0.70}  & \textbf{0.04}  & \textbf{5.30}  & \textbf{0.86}  & \textbf{1.03} & \textbf{0.26} & \textbf{12.6} & \textbf{1.99} 
 \\ \hline
 
 \textbf{Dataset:Dancing} 
    \\ 
 NCC \cite{hu2012quantitative} & 6.78 & 4.89 & 49.9 & 10.6  &  6.08   & 2.51  & 35.3 
 & 7.29           \\       
 Daisy \cite{tola2009daisy}  & 1.83 & 0.75 & 19.4 & 3.74 &  2.55   & 0.88  & 20.6  
 & 3.89            \\         
 MC-CNN \cite{zbontar2016stereo} & 1.12 & 0.38 & 13.8 & 2.52   &  1.76   & 0.39  & 17.3 & 3.41            \\
  Ours & \textbf{0.84}  & \textbf{0.16}  & \textbf{8.69}  & \textbf{1.68} & \textbf{1.71} & \textbf{0.33} & \textbf{15.3} & \textbf{3.01}  
  \\ \hline
 \\
\end{tabular}}
\caption{Depth estimation error results for 2 datasets against four compared methods are listed in the table. For details of experiment and error metrics, please refer to text.} 
\vspace{-0.5cm}
\label{tab:table1}
\end{table*}
\endgroup

\begingroup
\begin{table}[hb]
\resizebox{\columnwidth}{!}{
\begin{tabular}{lllll}
\\  \hline
  & \multicolumn{4}{c}{Lower is better}    \\ 
 \cline{2-5}  \\
    Method &  Abs Rel & Squ Rel & RMSE & $\text{RMSE}_{log}$
    \\    \hline
    \textbf{Dataset:Acting}
    \\
 MC-CNN \cite{zbontar2016stereo}   & 1.27 & 0.11 & 8.56 & 1.36 \\
 MC-CNN \cite{zbontar2016stereo} w/ constraint  & 0.76 & 0.08  & 7.30  & 1.16 \\
 Ours   & \textbf{0.70} & \textbf{0.04} & \textbf{5.30} & \textbf{0.86}  \\
 \\ \hline
  \textbf{Dataset:Dancing} 
    \\   
 MC-CNN \cite{zbontar2016stereo}  & 1.12 & 0.38 & 13.8 & 2.52 \\
 MC-CNN \cite{zbontar2016stereo} w/ constraint  & 0.98  & 0.31  & 12.3 & 2.41 \\
 Ours   & \textbf{0.84} & \textbf{0.16} & \textbf{8.69} & \textbf{1.68}    \\
\\
\end{tabular}}
\caption{Depth map evaluation with and without stereo constraint.} 
\vspace{-0.5cm}
\label{tab:table2_constraint}
\end{table}
\endgroup

\subsection{Results and Comparisons}


\begin{figure*}
  \includegraphics[width=\textwidth]{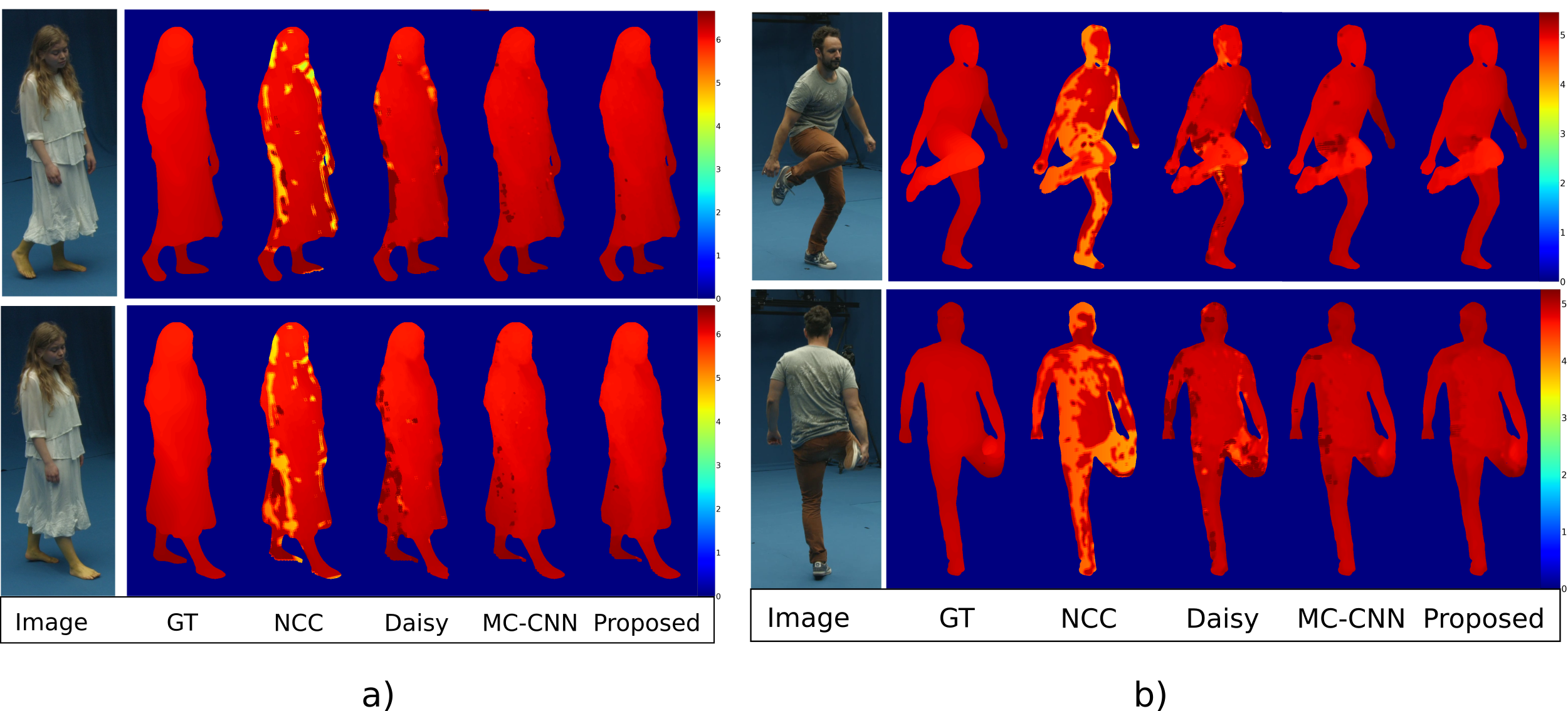}
  \caption{Comparison of estimated depth maps with ground-truth. Result depth maps of four methods, namely NCC \cite{hu2012quantitative}, Daisy \cite{tola2009daisy} and MC-CNN \cite{zbontar2016stereo}, including proposed one are illustrated. Camera baseline between stereo pairs are $24^{\circ}$. }
  \label{fig:Exp_depth_maps}
\end{figure*}

The proposed method is evaluated on a variety of real datasets with people in different environments, cluttered/controlled scene background, occlusions, camera settings and baselines: Acting, TV-Presenter, Dancing and Juggler \cite{CVSSP_3D}. The details of datasets are provided in Table \ref{tab:table_real_dataset}. These datasets consist of different dynamic human models and each scene is captured with number of cameras given in the Table. In these datasets, we use pseudo ground truth of 3D human reconstructions that are generated by using advanced multi-view camera capture system. For each camera view, ground-truth depth maps are rendered and then estimated 3D stereo reconstructions are evaluated against these rendered depth maps. Synthetic datasets for testing are different from training datasets and are generated using the framework explained in Section 3.2.

The proposed method is evaluated against baseline patch matching methods, namely NCC \cite{hu2012quantitative}, Daisy \cite{tola2009daisy} and MC-CNN \cite{zbontar2016stereo} since we propose a patch similarity based wide baseline stereo reconstruction method. MC-CNN is a state-of-the-art baseline method for stereo matching. MC-CNN is built on a Siamese network architecture and this network is trained on the Kitti \cite{Menze2015CVPRKitti} dataset of narrow baseline stereo street images taken from top of the car with sparse ground-truth obtained by lidar scanner.

We adopt the following error metrics \cite{eigen2014depth} to quantitatively evaluate the performance of our stereo reconstruction method. Established error metrics consider global statistics between a predicted depth map $d$ and its ground-truth depth image $d^{*}$ with $N$ depth pixels.  Specifically, we consider: ($i$) \textit{absolute relative error}: $\dfrac{1}{N}\sum_{i} \dfrac{\lvert{{d_i}-{{d_i}^*}}\rvert}{{d_i}^*}$; ($ii$) \textit{squared relative error}: $\dfrac{1}{N}\sum_{i} \dfrac{{\lvert\lvert{{d_i}-{{d_i}^*}}\rvert\rvert}^2}{{d_i}^*}$; ($iii$) \textit{root mean square error}: $\sqrt{\dfrac{1}{N}\sum_{i} ({{d_i}-{{d_i}^*}})^2}$; ($iv$) \textit{logarithmic root mean square error}: $\sqrt{\dfrac{1}{N}\sum_{i} ({{\log{d_i}}-{\log{d_i}^*}})^2}$. 

Table \ref{tab:table1} shows depth error metrics for two different datasets with two different wide baselines. For this experiment, baseline between stereo pairs is ${24^{\circ}}$ and ${36^{\circ}}$ for \textit{Acting}, ${22^{\circ}}$ and ${44^{\circ}}$ for \textit{Dancing} datasets. Corresponding depth estimation results are illustrated with ground-truth (GT) depth maps in Figure \ref{fig:Exp_depth_maps}. As shown in Table \ref{tab:table1}, the proposed method outperforms the baseline methods in terms of depth map estimation errors for wide baseline datasets. The proposed method gives approximately 25$\%$ RMSE error reduction for two camera baseline values compared to MC-CNN, which is the state of the art patch based stereo reconstruction method. It should also be considered that MC-CNN applies a series of expensive post processing steps, like occlusion removal, Semi-Global-Matching (SGM) \cite{hirschmuller2007stereo} and Bilateral filtering, where as proposed method only applies occlusion removal and not any of smoothing operations to recover wrong disparity estimations. Considering these post processing steps, for the same input stereo pairs with resolution of $3840x2160$ pixels, the run time for MC-CNN is 210 seconds whereas the proposed method only takes 135 seconds. Hence, the proposed method not only outperforms MC-CNN in depth error metrics, but also it is faster than MC-CNN by approximately 35$\%$.

\begin{figure}[t]
  \includegraphics[width=\columnwidth]{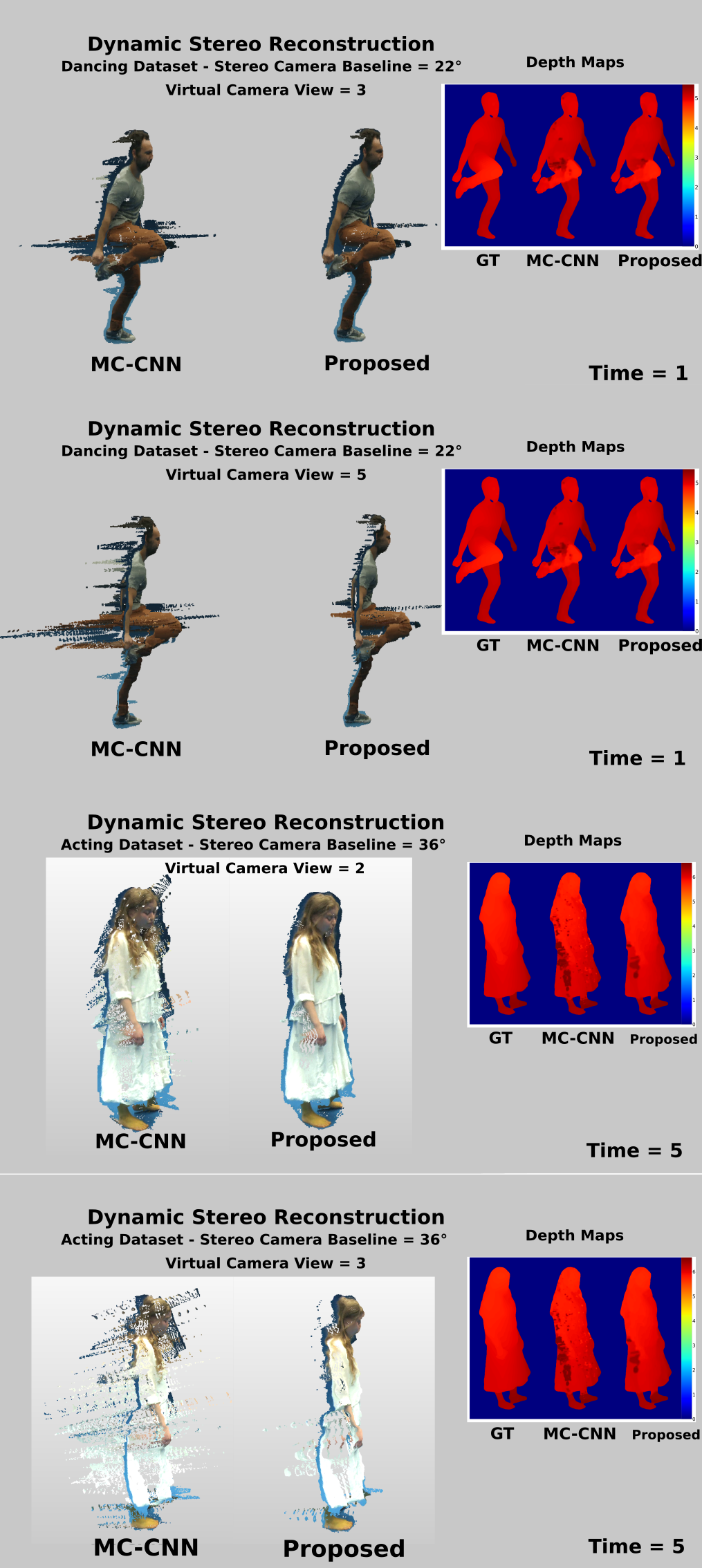}
  \caption{Point cloud stereo reconstruction results with depth map estimations from various time frames are illustrated for virtual camera views. \newline }
  \label{fig:Exp_point_clouds}
\end{figure}

Figure \ref{fig:Exp_point_clouds} shows the point clouds and depth maps, demonstrating a significant difference between the proposed method and MC-CNN. Depth values in the GT depth maps are defined in meters. Note that during the depth map error computation, only the foreground pixels are evaluated, and background pixels are discarded.

The proposed method also outperforms NCC \cite{hu2012quantitative} and Daisy \cite{tola2009daisy} in all depth estimation metrics. NCC \cite{hu2012quantitative} and Daisy \cite{tola2009daisy} generate local descriptors that are prone to fail in ambiguities, like repetitive textures, lack of textures, or lighting changes and large changes in shape. These failures can be resolved during post processing stage in wide baseline human stereo reconstruction methods \cite{richardt2016dense,leroy2017multi,mustafa2015general}. 

The reconstruction results are shown in Figure \ref{fig:Exp_point_clouds} with corresponding depth maps for MC-CNN and the proposed method. In addition to depth error metrics, 3D point clouds show the details in reconstruction. In Figure \ref{fig:Exp_point_clouds}, dynamic 3D stereo reconstruction of human body is also illustrated for different time frames. The generated point clouds are rendered to virtual cameras in order to see the stereo reconstruction errors that might be difficult to see from depth maps. The proposed method which learns from human specific features is able to capture details of clothing and hair which are challenging to reconstruct in wide baseline stereo setups. This answers the first question, that learning from a human-specific dataset improves wide baseline stereo performance.




Another contribution of our paper is to use semantic segmentation based stereo limitation to improve stereo matching performance or the reconstruction quality (Section \ref{M_ste_cons}). This constraint can be applied to any stereo matching method, so we evaluate the stereo matching performance of state-of-the-art methods with this constraint. During evaluation, only MC-CNN and the proposed stereo matching method are considered, because remaining methods' stereo reconstruction performance is not affected significantly with the constraint. In Table \ref{tab:table2_constraint}, semantic constrained is applied to MC-CNN for different datasets. Although semantic constraint increases the performance of MC-CNN by approximately 12$\%$ in RMSE, the proposed method still outperforms MC-CNN with stereo constraint in all error metrics, by average of $\%30$ in RMSE.



To evaluate the importance of the new \textit{S2P2} dataset, we evaluate performance of patch matching part of the proposed framework with two different models one of which is trained with \textit{S2P2} dataset, and other one is trained with Kitti dataset, shown in Table \ref{tab:table2_dataset}. Our method using the \textit{S2P2} trained network outperforms the network trained on Kitti, by approximately 30$\%$ in logarithmic RMSE. This basically shows that learning stereo matching from wide baseline and human specific data in our framework addresses more accurate wide baseline stereo reconstruction for people, which is the motivation of this paper. Table \ref{tab:table2_dataset} also demonstrates that data augmentation on stereo people dataset improves accuracy of depth maps and addresses the problem of domain shift from training on synthetic data and testing on real data.

\begingroup
\begin{table}[t]
\resizebox{\columnwidth}{!}{
\begin{tabular}{lllll}
\\  \hline
  & \multicolumn{4}{c}{Lower is better}    \\ 
 \cline{2-5}
    Method &  Abs Rel & Squ Rel & RMSE & $\text{RMSE}_{log}$
    \\    \hline
    \textbf{Dataset:Acting}
    \\
    Our Method w/ Kitti Dataset   & 0.85  & 0.12  & 8.7  & 1.41  \\
    Our Method w/ (${S2}{P2}$) no augmentation   & 0.67 & 0.09 & 7.8 & 1.25  \\
    Our Method w/ (${S2}{P2}$) Dataset  & \textbf{0.63} & \textbf{0.07} & \textbf{6.93} & \textbf{1.07} \\
    \hline

    \textbf{Dataset:Dancing}
    \\
 Our Method w/ Kitti Dataset   & 1.60  & 0.38 & 13.7 & 2.67  \\
 Our Method w/ (${S2}{P2}$) no augmentation   & 1.22 & 0.39 & 14.0 & 2.56  \\
 Our Method w/ (${S2}{P2}$) Dataset   & \textbf{0.81} & \textbf{0.11} & \textbf{7.68} & \textbf{1.52} \\
\end{tabular}}
\caption{Dataset and domain shift evaluation} 
\label{tab:table2_dataset}
\end{table}
\endgroup

As a part of our solution to scale variance in our method, we propose the pooling schema during inference stage of stereo reconstruction. In the pooling, we use patch size values of [9,19,35] in order to increase the patch scale variation. In order to show the effectiveness of the pooling stage, we evaluate proposed method with and without pooling and compare the results with MC-CNN \cite{zbontar2016stereo}. Since patch size is chosen as 9x9 in \cite{zbontar2016stereo}, we use this patch size during no-pooling evaluation. Depth estimation errors in Table \ref{tab:table2_scale_diversity} demonstrate that pooling stage in the pipeline increases accuracy of stereo reconstruction by solving scale diversity problem caused by domain shift. 

\begingroup
\begin{table}[b]
\resizebox{\columnwidth}{!}{
\begin{tabular}{lllll}
\\  \hline
  & \multicolumn{4}{c}{Lower is better}    \\ 
 \cline{2-5}
    Method &  Abs Rel & Squ Rel & RMSE & $\text{RMSE}_{log}$
    \\    \hline
    \textbf{Dataset:TV Presenter}, Patch Size = (9x9)
    \\
 MC-CNN \cite{zbontar2016stereo}   & 0.67 & 0.07 & 6.18 & 1.13 \\
 Our method + No Pooling & 0.61  & 0.06  & 5.94  & 1.08  \\
 Our Method + Pooling   & \textbf{0.60} & \textbf{0.05} & \textbf{5.33} & \textbf{0.98}  \\
\end{tabular}}
\caption{Scale Diversity Evaluation}
\label{tab:table2_scale_diversity}
\end{table}
\endgroup


\begin{figure}[t]
  \includegraphics[width=\columnwidth]{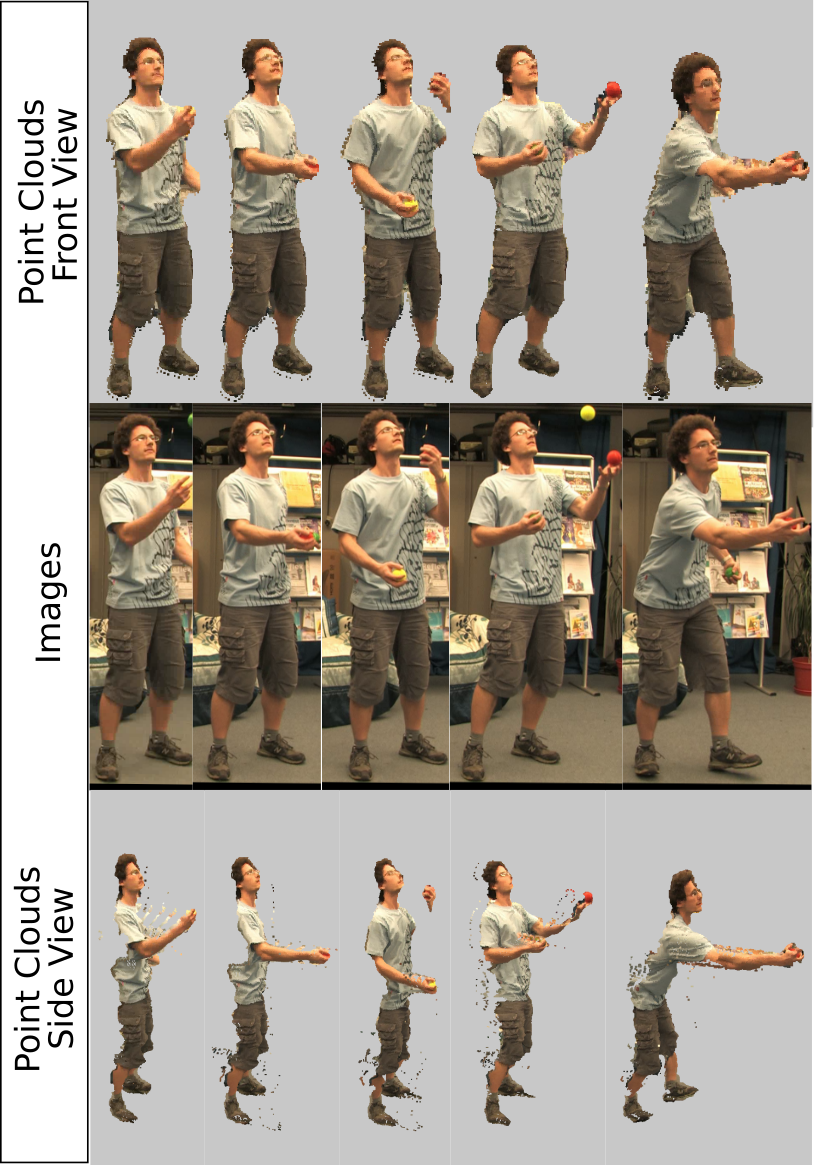}
  \caption{This figure illustrates the wide baseline stereo reconstruction of human in dynamic scene with cluttered background and ${{22}^{\circ}}$ camera baseline angle. \newline}
  \label{fig:Exp_dynamic_h_recons}
\end{figure}




To illustrate the performance of proposed method with human in dynamic scene with cluttered background, point cloud results are shown on \textit{Juggler} dataset for consecutive frames in Figure \ref{fig:Exp_dynamic_h_recons}. Note that stereo input images of juggler dataset are cropped for the visualization. The estimated point clouds from both front and side views show significant reconstruction performance from the proposed wide baseline stereo matching method with semantic human constraint. More stereo reconstruction results from both real and synthetic datasets are provided in supplementary files due to space constraint.


\section{Limitations}
The proposed method is developed for wide baseline stereo reconstruction for people, and this is not applicable to solve wide baseline stereo for generic scenes. However, a supervised learning based method for generic scenes is possible with provided training data and whole scene segmentation.

\section{Conclusion}
In this paper we proposed a method to solve the challenging task of wide baseline dense stereo reconstruction of humans. A framework to learn human specific features for stereo reconstruction from synthetic people stereo patch dataset is introduced. Multiple patch sizes are used to extract features and fused using pooling to address the problem of adapting the network from synthetic to real data. Comparative performance evaluation demonstrates that the learnt stereo matching outperforms state-of-the-art methods in human reconstruction and is robust to wide baseline and scale changes. To further refine the stereo reconstruction a person specific semantic stereo matching constraint is introduced. Extensive performance evaluation on real datasets shows that the proposed method outperforms state-of-the-art methods.

{\small
\bibliographystyle{ieee}
\bibliography{egbib}
}

\end{document}